\documentclass[11pt]{article}
\usepackage[
  letterpaper,
  margin=0.94in
]{geometry}
\usepackage{amsmath,amssymb,amsfonts}
\usepackage{natbib}
\usepackage{graphicx}
\usepackage{xcolor}
\usepackage[font=small]{caption}
\usepackage{subcaption}
\usepackage{mathtools}
\usepackage{multirow}
\usepackage{booktabs}
\usepackage{algorithm}
\usepackage{algorithmicx}
\usepackage{algpseudocode}
\algrenewcommand\algorithmicrequire{\textbf{Input:}}
\algrenewcommand\algorithmicensure{\textbf{Output:}}
\usepackage{tikz}
\usepackage{pgfplots}
\pgfplotsset{compat=1.18}
\usetikzlibrary{positioning, shapes, shadows, arrows, calc,
                 decorations.pathreplacing}
\pgfplotsset{
  every axis/.append style={
    axis line style={->},
    legend style={font=\scriptsize},
    label style={font=\scriptsize},
    title style={font=\scriptsize},
    tick label style={font=\scriptsize},
    axis x line*=bottom,
    axis y line*=left,
  }
}
\usepackage{listings}
\definecolor{codegreen}{HTML}{2E7D32}    
\definecolor{codepurple}{HTML}{7B1FA2}   
\definecolor{codeorange}{HTML}{E65100}   
\definecolor{codegray}{HTML}{757575}     
\definecolor{codebg}{HTML}{F5F5F5}       
\definecolor{codeframe}{HTML}{E0E0E0}    
\lstdefinestyle{manidreams}{
  language=Python,
  basicstyle=\ttfamily\scriptsize,
  keywordstyle=\color{codepurple}\bfseries,
  stringstyle=\color{codeorange},
  commentstyle=\color{codegray}\itshape,
  emph={SimulationBasedTSIP, LearnedBasedTSIP,
        DRISCage, PolicySampler, ManiDreamsEnv,
        PushCubeEnv, MPPIOptimizer, CageController},
  emphstyle=\color{codegreen}\bfseries,
  numbers=left,
  numberstyle=\tiny\color{codegray},
  numbersep=6pt,
  backgroundcolor=\color{codebg},
  frame=single,
  rulecolor=\color{codeframe},
  framesep=4pt,
  xleftmargin=12pt,
  captionpos=b,
  framexleftmargin=12pt,
  tabsize=4,
  breaklines=true,
  showstringspaces=false,
  columns=flexible,
  aboveskip=8pt,
  belowskip=4pt,
  morekeywords={self},
  float=t,
  floatplacement=t,
}
\usepackage{textcomp}
\usepackage{csquotes}
\usepackage{todonotes}
\usepackage{authblk}
\usepackage[
  colorlinks=true,
  linkcolor=blue!70!black,
  citecolor=green!50!black,
  urlcolor=blue!70!black
]{hyperref}
\newcommand{\api}[1]{\textsf{#1}}          
\newif\ifanonymous
\anonymousfalse  
\title{\Large \textbf{ManiDreams: An Open-Source Library for Robust Object
Manipulation via Uncertainty-aware Task-specific Intuitive Physics}}

\author[1]{Gaotian Wang}
\author[1]{Kejia Ren}
\author[2]{Andrew S.\ Morgan}
\author[1]{Kaiyu Hang}
\affil[1]{Department of Computer Science, Rice University, Houston, TX 77005, USA}
\affil[2]{Robotics and AI Institute, Cambridge, MA 02142, USA}
\date{}

\begin{document}
\maketitle

\begin{abstract}

Dynamics models, whether simulators or learned world models, have long been central to robotic manipulation, but most of these models focus on minimizing prediction error rather than confronting a more fundamental challenge: real-world manipulation is \emph{inherently} uncertain. We argue that robust manipulation under uncertainty is fundamentally an \emph{integration problem}: uncertainties must be represented, propagated, and constrained within the planning loop, not merely suppressed during training.

We present and open-source \textit{ManiDreams}, a modular framework for uncertainty-aware manipulation planning over intuitive physics models that realizes this integration through composable abstractions for distributional state representation, backend-agnostic dynamics prediction, and declarative constraint specification for action optimization. The framework explicitly addresses three sources of uncertainty: \emph{perceptual}, \emph{parametric}, and \emph{structural}. It wraps any base policy with a sample-predict-constrain loop that evaluates candidate actions against distributional outcomes, adding robustness without retraining. Experiments on default ManiSkill tasks show that ManiDreams maintains robust performance under various perturbations, where the RL baseline degrades significantly.
Runnable examples on pushing, picking, catching, and real world deployment demonstrate flexibility for applications across different policies, optimizers, physics backends, and executors.
The framework is publicly available at: \url{https://github.com/Rice-RobotPI-Lab/ManiDreams}

\end{abstract}

\begin{figure*}[t]
  \centering
  \includegraphics[width=1\columnwidth]{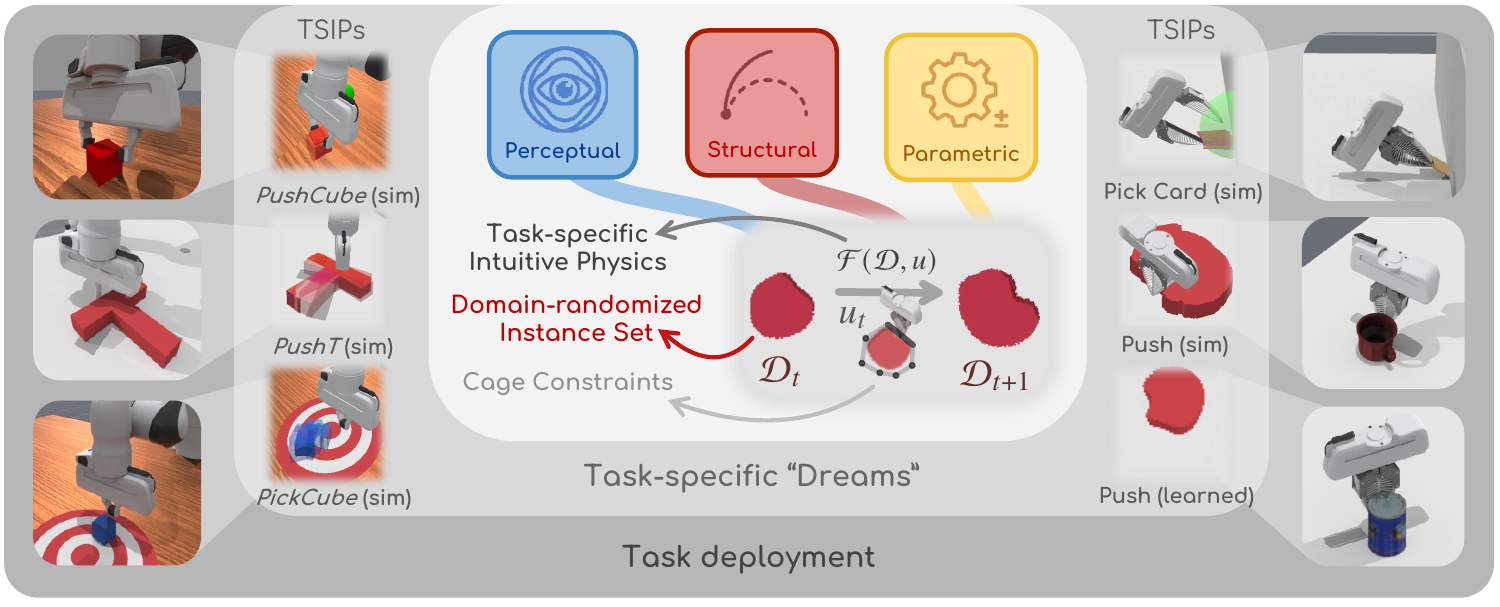}
  \caption{ManiDreams unifies perceptual, parametric, and structural uncertainties through Task-specific Intuitive Physics (TSIP) and Domain-randomized Instance Set (DRIS).
  \textit{Left}: Simulation-based TSIP across standard tasks (\textit{PushCube}, \textit{PickCube}, \textit{PushT}), where the overlaid objects are DRIS instances.
  \textit{Right}: A card picking task and a pushing task driven by simulation-based and learning-based TSIPs.}
  \label{fig:splash}
  \vspace{-10pt}
\end{figure*}

\section{Introduction}
\label{sec:intro}

Dynamics models, whether physics simulators or learned world models, serve as cornerstones of robotic manipulation~\cite{reviewdynamics}. Although most models focus on minimizing prediction error, they neglect a more fundamental challenge: real-world manipulation is \emph{inherently} uncertain.
Three sources of uncertainty arise jointly during execution: \emph{perceptual} uncertainty from noisy and delayed observations, \emph{parametric} uncertainty from inaccurate physical parameter estimation, and \emph{structural} uncertainty from imperfect dynamics models~\cite{muratore2022robot}.
Existing pipelines largely treat these as training-time concerns.
Domain randomization~\cite{tobin2017domain, peng2018sim} injects variability during training, and learned world models~\cite{hafner2023dreamerv3, hansen2024tdmpc2} absorb some mismatch into latent representations, but once a policy is deployed, the explicit uncertainty structure is discarded~\cite{wu2022daydreamer}.

Recent findings~\cite{wang2025cagingintime} suggest that 
robust manipulation under uncertainty is fundamentally a \emph{unification} problem: when perceptual, parametric, and structural uncertainties are handled independently, their errors accumulate throughout the pipeline, calling for a unified framework that represents, propagates, and constrains them jointly. Today's tools address only fragments of this challenge. Simulation frameworks~\cite{tao2024maniskill3} support domain randomization but still propagate only a single object instance per environment. World models, from physics engines to diffusion-based predictors~\cite{yang2024diffusion}, produce single-state forecasts with no mechanism to represent or propagate uncertainty across plausible outcomes. 
Notably, the pattern of sampling candidate actions, predicting their outcomes through a world model, and filtering by a constraint criterion is emerging independently across recent systems~\cite{cosmospolicy, hansen2024tdmpc2}, yet no shared, composable framework exists for this paradigm with uncertainties considered.

We present \textbf{\textit{ManiDreams}}, an open-source modular Python framework for uncertainty-aware manipulation planning that closes these gaps through composable abstractions. As shown in Fig.~\ref{fig:splash}, \emph{Domain-Randomized Instance Set} (DRIS) overlays multiple object instances with randomized physical parameters into a unified state, capturing perceptual and parametric uncertainty through diverse pose hypotheses and varied physical properties. \emph{Task-Specific Intuitive Physics} (TSIP) interface propagates the DRIS forward under candidate actions. Because TSIP exposes a standardized interface, any dynamics backend (simulator or learned model) can serve as the propagator, accommodating structural uncertainty from model mismatch. \emph{Caging constraints} then bound the growing uncertainty, providing explicit boundaries that a \emph{Solver} enforces through constraint-driven optimization. All four abstractions follow a plugin architecture, making ManiDreams a general realization of the \emph{sample-predict-constrain} paradigm. ManiDreams ships with ManiSkill integration, multiple TSIP backends, pre-built caging constraints, multiple solvers, and a detailed documentation page.

We showcase ManiDreams through extensive demonstrations and evaluations. Runnable examples on pushing, picking, and catching validate composability across different TSIP backends, caging constraints, and solvers. Real-robot deployment on cluttered picking tasks demonstrates zero-shot sim-to-real transfer. Robustness benchmarks on three ManiSkill tasks under escalating perturbations quantify the advantage over a standard PPO baseline. Ablation studies and runtime profiling further characterize the framework's parameter sensitivity and computational overhead.

In summary, this paper makes three core contributions:
\begin{enumerate}
  \item A composable, open-source framework realizing the \emph{sample-predict-constrain} paradigm through plugin-based abstractions.
  \item A unified distributional dynamics (DRIS + TSIP + cages) that represents, propagates and constrains uncertainties via a standardized interface.
  \item Comprehensive evaluations spanning runnable examples, benchmarks under all three uncertainty types, runtime analysis, and real-robot deployment.
\end{enumerate}
\section{Related Work}
\label{sec:related}

\emph{World Models for Manipulation.}
GPU-parallelized simulators have converged on well-engineered, unified APIs that train RL policies at thousands of FPS with standardized task definitions and reward interfaces~\cite{makoviychuk2021isaacgym, tao2024maniskill3}. Learned world models complement these simulators by planning in compact latent spaces~\cite{hafner2023dreamerv3, hansen2024tdmpc2}, extending the generative models to physical robots~\cite{wu2022daydreamer, alonso2024diffusion, ding2024diffusionwm}. However, both families produce single-state forecasts without exposing uncertainty to planners. Furthermore, each model ships with its own state representation and rollout protocol; no unified platform wraps heterogeneous dynamics backends behind a
common, uncertainty-aware prediction interface.

\emph{Domain Randomization and Uncertainty.}
Domain randomization~\cite{tobin2017domain, peng2018sim} injects variability during training to bridge the sim-to-real gap, with follow-up work automating the randomization distribution~\cite{chebotar2018closing, mehta2019active, muratore2022robot}. Some learned world models encode uncertainty signals internally, such as ensemble disagreement, latent-space variance, and sample diversity~\cite{yang2024diffusion}, but these signals are consumed for exploration bonuses or model selection rather than exposed as first-class distributional outputs. No existing framework propagates uncertainty as an explicit distributional state that a planner can constrain at execution time.

\emph{Constraints in Manipulation.}
Control barrier functions~\cite{oscbf2025} enforce forward invariance via real-time quadratic-program filters~\cite{nudge2026}, and robust model-predictive control tightens constraints against bounded disturbances using tube-based or min-max formulations~\cite{mayne2005robustmpc}. These methods are effective, but assume known, low-dimensional dynamics, leaving open how to constrain distributional uncertainty over longer horizons. On task level, caging provides topological confinement guarantees regardless of exact object state~\cite{rimon1999caging, rodriguez2012caging, makita2017survey}, recently extended to guarantee containment along entire manipulation horizons under physical and perceptual uncertainties~\cite{dong2024codesigningtoolscontrolpolicies, wang2025cagingintime}. However, every existing caging implementation is tightly coupled to a specific task geometry~\cite{cageinmotion}; a reusable declarative constraint library analogous to OMPL~\cite{sucan2012ompl} remains absent.

\emph{Sample-Predict-Constrain Pattern.} 
A recurring pattern across planning systems is to sample candidate actions, predict outcomes through a dynamics model, and filter by a cost or constraint criterion. This pattern appears in sampling-based planners~\cite{MPPI, pmlr-v155-pinneri21a}, latent-space model-predictive controllers~\cite{hansen2024tdmpc2}, and recent world-model-based policy optimization systems that pair generative video prediction with value-function ranking~\cite{cosmospolicy}. Despite this convergence, each system couples the dynamics model, cost function, and action selection monolithically; the components cannot be independently replaced or reused across tasks. ManiDreams abstracts this recurring pattern into a composable framework with independently replaceable components.
\section{System Architecture}
\label{sec:architecture}

This section describes the static structure of ManiDreams, organized around four composable abstractions that map directly to the \emph{sample-predict-constrain} paradigm: Domain-Randomized Instance Set (DRIS) for distributional state representation, Task-Specific Intuitive Physics (TSIP) for forward dynamics prediction, caging constraints for bounding uncertainty, and solvers for action selection.

\subsection{Design Principles}
\label{sec:principles}

We follow three core principles in building ManiDreams, ensuring that uncertainty handling remains modular, portable, and easy to adopt.

\textit{Distributional-first.} Every DRIS treats a set of possible states, rather than a single point, as the basic unit of reasoning; uncertainty propagates through all pipeline stages, which is the core contribution that sets ManiDreams apart from other frameworks with dynamics models.

\textit{Backend-agnostic.} All dynamics-dependent logic is isolated behind a single forward-prediction interface, so the same constraints and solvers work unchanged whether the backend is a GPU simulator or a learned world model.

\textit{Minimal intrusion.} ManiDreams wraps any Gym-compatible environment with a thin adapter; switching between direct policy execution and uncertainty-aware planning requires changing a single argument.

\subsection{Core Abstractions}
\label{sec:abstractions}

\begin{figure}[t]
  \centering
  \includegraphics[width=0.5\columnwidth]{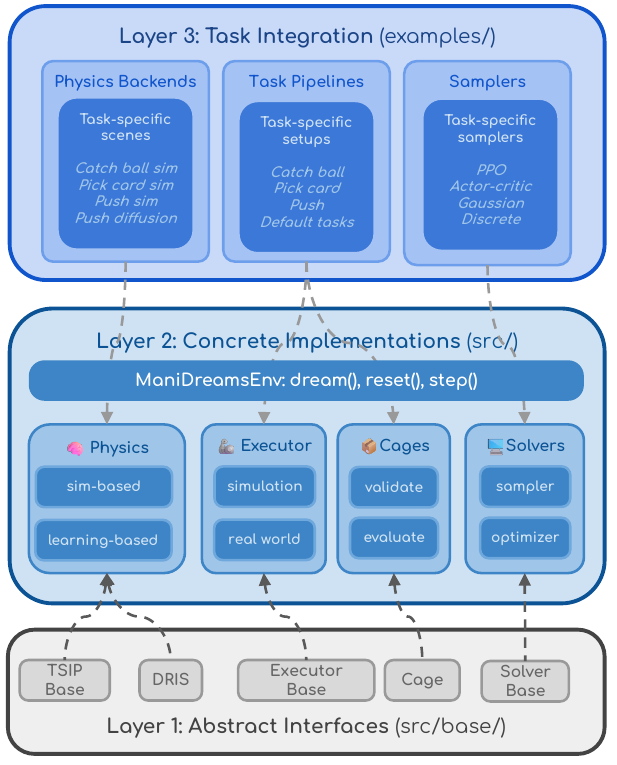}
  \caption{Three-layer architecture. \textit{Layer~1} defines abstract interfaces for the four core abstractions, analogous to OMPL's \textsf{StateSpace}, \textsf{StatePropagator}, \textsf{StateValidityChecker}, and \textsf{Planner}. \textit{Layer~2} provides concrete implementations unified behind a single environment class. \textit{Layer~3} is user-facing: adding a new task requires only populating this layer. Dashed arrows indicate inheritance.}
  \label{fig:architecture}
  \vspace{-15pt}
\end{figure}

Fig.~\ref{fig:architecture} shows the three-layer organization, following the pattern established by libraries such as OMPL~\cite{sucan2012ompl}. Layer~1 defines abstract interfaces for the four core abstractions. Layer~2 provides concrete implementations unified behind a single environment class. Layer~3 is user-facing: adding a new task requires only supplying a scene description, a pipeline configuration, and an action sampler; no framework code needs modification. Sec.~\ref{sec:pipeline} details how these abstractions compose at runtime.

We first introduce the formal definitions of DRIS and TSIP that underlie the framework. Consider a Contextual Markov Decision Process $(\mathcal{S}, \mathcal{U}, \mathcal{C}, P)$, where $\mathcal{S}$ is the object state space, $\mathcal{U}$ the robot action space, $\mathcal{C}$ the context space capturing physical variations (shape, mass, friction, etc.), and $P(s' \mid s, u, c)$ the transition probability. Given a fixed context $c \in \mathcal{C}$, we assume the transition is approximately deterministic: $s_{t+1} = f(s_t, u_t, c)$.

\subsubsection{State Representation (DRIS)}
\label{sec:abs_dris}

DRIS $\mathcal{D}$ is the universal state representation that flows through all pipeline stages. Rather than carrying a single point estimate, each DRIS bundles the observation with a distributional context that captures the collective physical futures of the target object under parameter uncertainty. Concretely, we sample a discrete set $\hat{\mathcal{C}} = \{c^{(i)}\}_{i=1}^{m}$ from the context space and define the DRIS at time $t$ as the collection of state-context pairs:
\begin{equation}
\label{eq:dris}
  \mathcal{D}_t
  \;=\;
  \bigl\{\bigl(s_t^{(i)},\, c^{(i)}\bigr)
        \;\bigm|\;
        c^{(i)} \in \hat{\mathcal{C}}
  \bigr\}_{i=1}^{m}
  \;\subset\; \mathcal{S} \times \mathcal{C},
\end{equation}
where $s_t^{(i)} \in \mathcal{S}$ is the state of the $i$-th instance evolving under its context $c^{(i)}$. The state projection $\mathcal{S}_t = \mathrm{proj}_{\mathcal{S}}(\mathcal{D}_t) = \{s_t^{(i)}\}_{i=1}^{m}$ extracts the set of all instance states, where distributional statistics are computed. DRIS is modality-agnostic: the observation field can hold a state vector, an RGB image, or a point cloud, while the distributional context format remains uniform. 

\subsubsection{Forward Dynamics (TSIP)}
\label{sec:abs_tsip}

TSIP $\mathcal{F}$ provides the forward-prediction interface that connects DRIS to the constraint layer. Depending on the backend, it either applies the state transition $f$ to every instance simultaneously (simulation-based) or operates on the entire DRIS as a single input (learning-based):
\begin{equation}
\label{eq:tsip}
\begin{aligned}
  \mathcal{D}_{t+1}
  &= \mathcal{F}(\mathcal{D}_t,\, u_t) \\
  &= \bigl\{\bigl(f(s_t^{(i)}, u_t, c^{(i)}),\,
         c^{(i)}\bigr)
        \bigm|
        (s_t^{(i)}, c^{(i)}) \in \mathcal{D}_t
  \bigr\}
\end{aligned}
\end{equation}
This formulation converts the stochastic dynamics over individual contexts into a deterministic mapping over distributions: uncertainty is captured in the representation rather than the dynamics model. Given the current DRIS $\mathcal{D}_t$ and an action (or a batch of $N$ candidates), \api{TSIP.next()} returns the predicted next DRIS following Eq.~\eqref{eq:tsip}.

In this work, we provide two backends: a \emph{simulation-based} backend that wraps ManiSkill3~\cite{tao2024maniskill3} and evaluates all candidates with their $m$ DRIS copies in parallel on GPU, and a \emph{learning-based} backend that wraps a diffusion model operating in a feature space $\mathbf{z}_t = \psi(\mathcal{S}_t) \in \mathbb{R}^{d_z}$. Both produce the same DRIS output format, keeping all downstream modules unchanged. Users may add further backends by implementing the \api{next()} interface.

\subsubsection{Caging Constraint}
\label{sec:abs_cage}
Caging constraints bound the growing uncertainty of a DRIS to maintain task-level safety, requiring that $\mathcal{S}_{t+1} \subseteq \mathcal{S}_{\mathrm{cage}}^{t+1}$. Each constraint exposes \api{evaluate()} (continuous cost) and \api{validate()} (binary containment flag). Unlike classical state-validity checkers that reason about a single configuration, caging constraints reason about the spread of the distribution. Built-in variants include geometric, trajectory-based, plate-based, pixel-space, and composite constraints; users define custom ones by subclassing and implementing the two methods.

\subsubsection{Solver}
\label{sec:abs_solver}
The Solver closes the sample-predict-constrain loop by combining a \emph{sampler} that proposes $N$ candidate actions (e.g., a policy's stochastic head, a Gaussian, or a discrete set) with an \emph{optimizer} that scores candidates via the caging constraint and returns the best valid action (e.g., $N$-best selection or MPPI). Any sampler can be paired with any optimizer, adding a safety filter on top of existing policies without retraining.
\begin{figure*}[t]
  \centering
  \includegraphics[width=1\columnwidth]{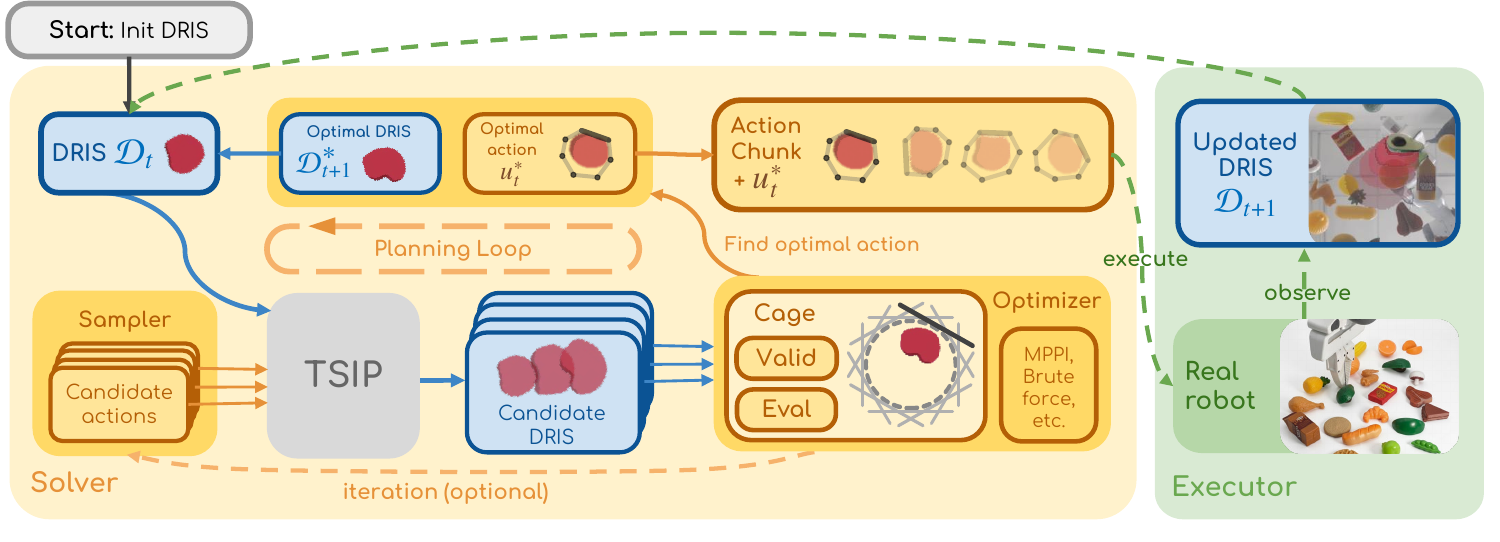}
  \caption{Runtime pipeline of ManiDreams. The \textbf{planning loop} (yellow): The solver generates candidate actions, predicts their DRIS outcomes via TSIP, and selects the best action satisfying the caging constraint; the loop may iterate for refinement. The \textbf{execution loop} (green): The selected action is appended to an action chunk; once complete, the chunk is dispatched to the real robot or simulator, which observes the outcome and produces an updated DRIS for the next planning cycle.}
  \label{fig:pipeline}
\end{figure*}

\section{Framework Pipeline}
\label{sec:pipeline}

This section describes how the abstractions defined in Sec.~\ref{sec:architecture} compose at runtime to realize the sample-predict-constrain paradigm.

\subsection{Two-Level Execution Loop}
\label{sec:loop}

Fig.~\ref{fig:pipeline} shows the full runtime pipeline, which consists of two nested loops for planning and execution.

\textit{Planning.} Given the current DRIS~$\mathcal{D}_t$, the solver performs the cycle shown in Algorithm~\ref{alg:cit}: a sampler proposes $N$ candidate actions (line~4); \api{TSIP.next()} predicts the resulting DRIS for each candidate in a single batched call (line~5); \api{evaluate()} and \api{validate()} score and filter every predicted outcome; and the optimizer returns the best valid action (line~6). If the optimizer supports iterative refinement (e.g., MPPI warm-starting), the loop repeats from line~4 with a refined proposal. When $N{=}1$ or no constraint is active, the pipeline reduces to direct policy execution, recovering standard baseline behavior.

\textit{Execution.}  After each planning loop, the selected action is appended to an action chunk. Once the chunk reaches the configured length, it is dispatched to the real robot or simulation executor. After execution, the executor observes the resulting state and constructs an updated DRIS~$\mathcal{D}_{t+1}$, which becomes the input for the next planning cycle. This outer loop runs at the physical control frequency and is independent of the planning model, enabling sim-to-real transfer by simply swapping the executor.

\begin{algorithm}[t]
\caption{Action Selection under Caging Constraints}\label{alg:cit}
\begin{algorithmic}[1]
\Require TSIP with $N$ eval envs ($m$ DRIS copies each), Cage $\mathcal{G}$, sampler $\pi$, timestep $t$
\State $\mathcal{G}.\Call{Update}{t}$ \Comment{Update cage geometry}
\State Sync executor state $\to$ all $N$ TSIP envs
\State $\mathcal{D}_t \gets \Call{TSIP.GetDRIS}{\,}$ \Comment{Observe}
\State $\{u_i\}_{i=1}^{N} \gets \pi(\mathcal{D}_t)$ \Comment{Sample candidate actions}
\State $\{\hat{\mathcal{D}}_{t+1}^{(i)}\} \gets \Call{TSIP.Next}{\mathcal{D}_t,\;\{u_i\}}$ \Comment{Propagate}
\State $\begin{cases} c_i \gets \mathcal{G}.\Call{Evaluate}{\hat{\mathcal{D}}_{t+1}^{(i)}} \\ v_i \gets \mathcal{G}.\Call{Validate}{\hat{\mathcal{D}}_{t+1}^{(i)}} \end{cases}$ \Comment{Caging constraints}
\State $i^{*} \gets \arg\min_{i:\,v_i=1}\; c_i$ \Comment{Find optimal action}
\State \textbf{return} $u_{i^{*}},\;\hat{\mathcal{D}}_{t+1}^{(i^{*})}$
\end{algorithmic}
\end{algorithm}
\vspace{-2pt}

\subsection{DRIS Propagation}
\label{sec:propagation}

At each call to \api{TSIP.next()}, the state-sync stage (line~2 of Algorithm~\ref{alg:cit}) broadcasts the executor's current target pose to all $N$ evaluation environments, where each environment resets its $m$ DRIS copies with fresh perturbations drawn from a user-configured distribution. All $N{\times}m$ instances are then stepped under the candidate actions. Because copies start from slightly different poses and may encounter different contact conditions, their post-step positions naturally diverge; the backend aggregates these into per-candidate distributional statistics (mean and per-axis variance), stored in the returned DRIS. The downstream cage and solver consume these statistics identically regardless of whether the backend is simulation-based or learning-based.

\subsection{Operational Modes}
\label{sec:modes}

The loop separation supports two modes. In \emph{online mode}, the inner loop runs once per control step and the selected action is immediately dispatched, suiting reactive tasks such as catching. In \emph{plan-then-execute mode}, the inner loop iterates for a full horizon~$T$ via \api{dream(horizon)}, producing a complete action sequence that the executor replays independently. Because planning and execution do not share state, the planning model can use simplified dynamics while the executor runs full-fidelity simulation or real hardware.
\section{Implementation Details}
\label{sec:implementation}

While Sec.~\ref{sec:architecture} described the conceptual abstractions, this section focuses on engineering details for extending the framework and using its API. ManiDreams is distributed as a Python package requiring only PyTorch and ManiSkill3 as core dependencies.

\subsection{Backend Integration}
\label{sec:backend}

Each of the four core abstractions can be extended by subclassing and implementing a small set of methods. DRIS is configured by specifying randomization ranges for physical parameters (mass, friction, geometry) and the instance count $m$; at each \api{reset()}, the backend samples $m$ contexts and overlays them into a single scene to form $\mathcal{D}_0$. ManiDreams provides two main TSIP implementations: \api{SimulationBasedTSIP} wraps ManiSkill3 with no offline training required, while \api{LearningBasedTSIP} loads a pre-trained forward dynamics model (e.g., a diffusion world model) for a specific task. New backends subclass \api{TSIPBase} and implement \api{next(dris, action)} and \api{reset()}.

Custom caging constraints subclass \api{Cage} and implement \api{evaluate(dris)} and \api{validate(dris)}. Executors subclass \api{ExecutorBase} and implement \api{execute(action)} and \api{get\_obs()}. Because the executor maintains its own environment instance independent of the TSIP, swapping from simulation to real-world execution requires changing only the executor class. Solvers subclass \api{SolverBase} and implement \api{solve()}; the framework ships with both sampling (e.g., \api{PolicySampler}) and optimization (e.g., \api{MPPIOptimizer}) components.

\subsection{User-Facing API}
\label{sec:api}

\api{ManiDreamsEnv} is the single entry point for all interactions. It wraps any Gym-compatible environment and exposes three methods: \api{reset()} for episode initialization, \api{step(action)} for standard Gym-compatible stepping, and \api{dream(horizon)} for running the full sample-predict-constrain loop over a planning horizon~$T$.

\begin{figure}[t]
  \centering
  \includegraphics[width=0.6\columnwidth]{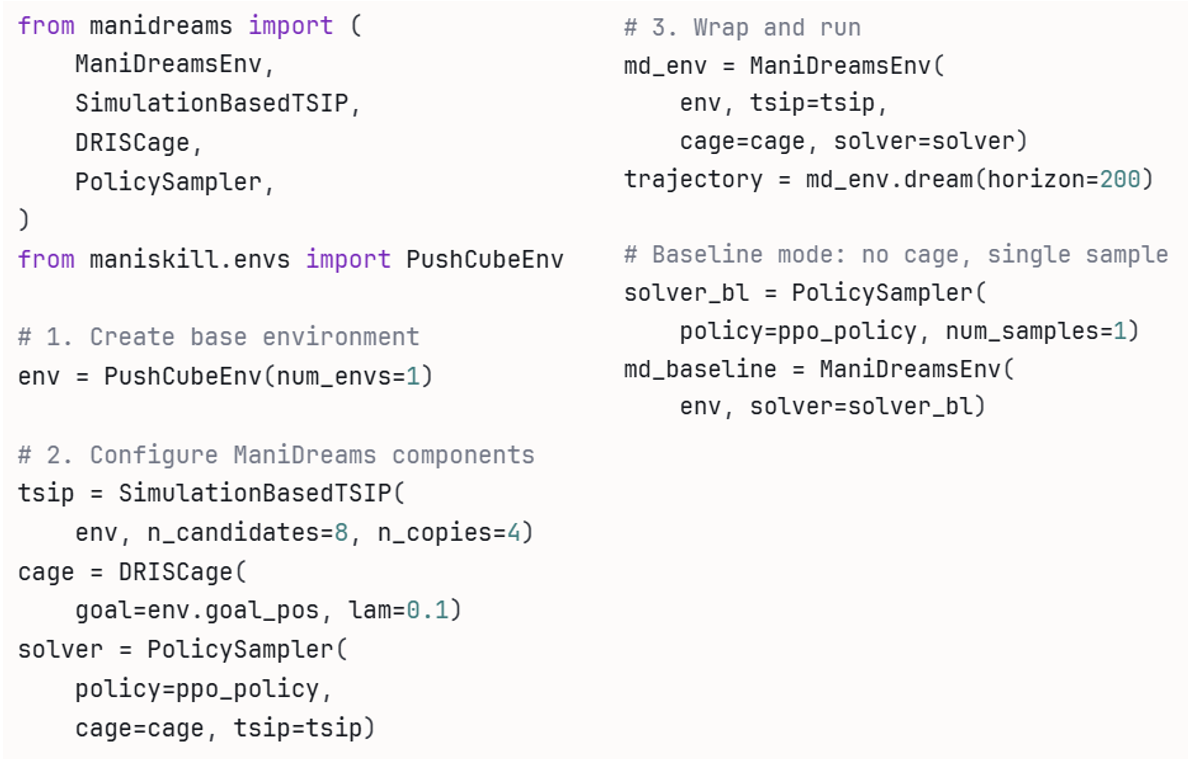}
  \caption{Minimal working example. Three steps configure and launch the full sample-predict-constrain pipeline. Setting \api{num\_samples=1} and omitting the cage reduces the pipeline to direct policy execution, serving as the baseline in all experiments.}
  \label{fig:code}
  \vspace{-15pt}
\end{figure}

Fig.~\ref{fig:code} illustrates a minimal working example: the import block and step~1 create the base environment; step~2 configures the four abstractions (TSIP, cage, solver with a PPO policy); and \api{dream(horizon)} in step~3 runs the complete planning loop. Setting \api{num\_samples=1} and omitting the cage causes the solver to bypass TSIP evaluation entirely, reproducing the standard RL deployment loop. This ensures that all comparisons in Sec.~\ref{sec:experiments} are conducted under identical conditions except for the presence of the sample-predict-constrain loop.

\section{Experiments}
\label{sec:experiments}

We evaluate ManiDreams on both its software design (modularity, ease of implementation, lightweight integration) and its effectiveness for robust manipulation under uncertainty with five sets of experiments: runnable examples illustrating composability across tasks and TSIPs (Sec.~\ref{sec:exp_qualitative}), real-world deployment on high-uncertainty tasks (Sec.~\ref{sec:exp_real}), robustness benchmarks against an RL baseline under escalating perturbations (Sec.~\ref{sec:exp_robust}), ablation studies on key framework parameters (Sec.~\ref{sec:ablation}), and runtime overhead analysis (Sec.~\ref{sec:exp_overhead}).

All simulations were built on ManiSkill3~\cite{tao2024maniskill3}. The learning-based TSIP loaded a diffusion model checkpoint trained with simulation rollouts. Real-world experiments used a Franka Panda arm equipped with Finray soft grippers. All computation, including TSIP propagation and solver optimization, ran on a single NVIDIA RTX 5070 Ti laptop GPU.

\subsection{Runnable Examples}
\label{sec:exp_qualitative}

We first present three reference examples in random object pushing, card picking and ball catching, to demonstrate the framework's flexibility, each highlighting a different axis of composability. All three examples shared the same \api{ManiDreamsEnv} entry point and differed only in which Layer~2 components were plugged in, validating the composability claimed in Sec.~\ref{sec:architecture}.

\begin{figure}[t]
  \centering
  \includegraphics[width=0.6\columnwidth]{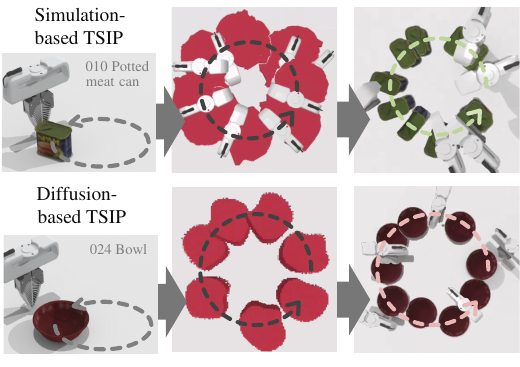}
  \caption{Object pushing with circular trajectory following using different TSIPs. Random target objects are selected from the YCB dataset~\cite{calli2017ycb}. \textit{Top}: Simulation-based TSIP. \textit{Bottom}: Learning-based TSIP based on a pre-trained diffusion model. }
  \label{fig:pushing}
  \vspace{-15pt}
\end{figure}

\textit{Pushing (TSIP flexibility).}
As shown in Fig.~\ref{fig:pushing}, the robot pushed a randomly selected YCB object~\cite{calli2017ycb} along a circular reference trajectory.
We ran this task with two TSIPs: \api{SimulationBasedTSIP} and \api{LearningBasedTSIP}, using identical cage, sampler (discrete action sampling), and solver (n-best selection) configurations. Both TSIPs produced successful trajectories, confirming that swapping the TSIP requires no changes to the Cage or Solver.

\begin{figure}[t]
  \centering
  \includegraphics[width=0.6\columnwidth]{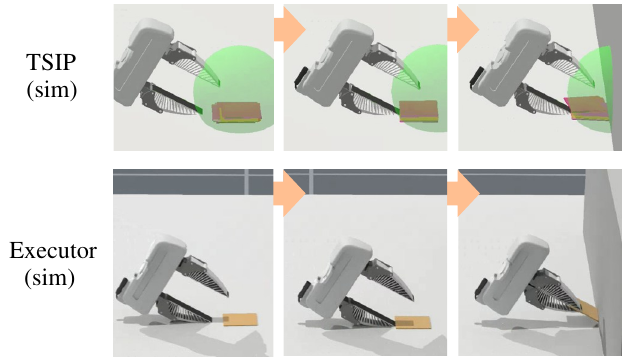}
  \caption{Card picking example with a simulation-based TSIP and a \api{Geometric3DCage} (transparent green sphere).}
  \label{fig:picking}
\end{figure}

\textit{Card Picking (Cage and solver flexibility).}
Beyond 2D pushing, we deployed ManiDreams on a 3D card picking task shown in Fig.~\ref{fig:picking}, where the robot picked a thin card by pushing it to the corner. This task used a \api{Geometric3DCage} that tracked a pre-set 3D waypoint trajectory, a continuous random sampler over a 3D action space $(x, y, \text{yaw})$, and an MPPI solver for iterative refinement, confirming that each abstraction is independently replaceable.

\begin{figure}[t]
  \centering
  \includegraphics[width=0.6\columnwidth]{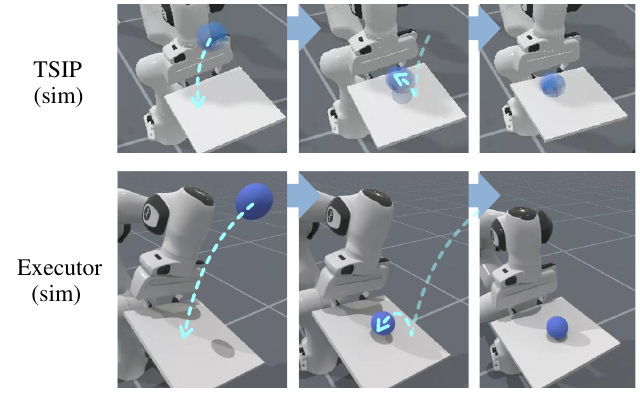}
  \caption{Dynamic ball catching example with a simulation-based TSIP and a pretrained PPO sampler.}
  \label{fig:catching}
\end{figure}

\textit{Ball Catching (Sampler flexibility).}
As shown in Fig.~\ref{fig:catching}, the robot caught a ballistic object launched toward a plate. 
A pretrained PPO policy served as the sampler with $N$ candidate actions from its stochastic action head, each evaluated in a separate DRIS-augmented environment via simulation-based TSIP, and the solver selected the best valid action, confirming that the sampler is independently replaceable.

\subsection{Real Robot Deployment}
\label{sec:exp_real}

\begin{figure*}[t]
  \centering
  \includegraphics[width=1\columnwidth]{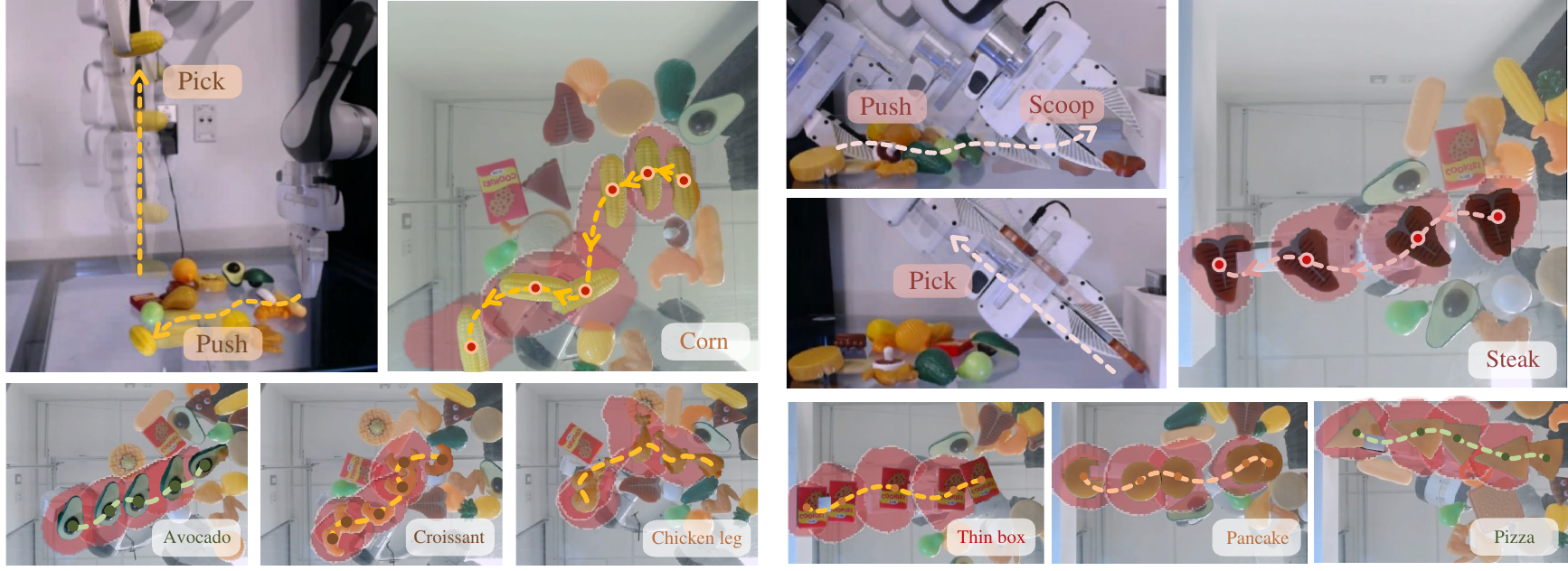}
  \caption{Real-robot deployment. \textit{Left}: Cluttered push-picking (corn, avocado, chicken leg, croissant), where the robot pushes obstructing objects aside before grasping the target. \textit{Right}: Scooping via wall corner for flat objects (steak, pancake, thin box, pizza). In both tasks, ManiDreams plans with learning-based TSIP and pixel-based DRIS (red region behind the object) for consecutive action chunks.}
  \label{fig:realrobot}
\end{figure*}

We instantiated the pushing and card picking examples from Sec.~\ref{sec:exp_qualitative} zero-shot on a real Franka Panda arm by swapping the executor to \api{RealWorldExecutor}. As shown in Fig.~\ref{fig:realrobot}, both tasks used diffusion-based TSIP and were deployed in cluttered picking scenarios with much greater uncertainty than simulation, targeting regular objects (pushing) and plate-shaped objects (card picking) respectively.

An RGB camera mounted beneath a transparent tabletop captured a bottom-up view, from which SAM2~\cite{ravi2024sam2} segmented the target object on the $xy$ plane directly as the pixel-based DRIS input. No additional object modeling was required beyond this segmentation. Execution proceeded as a sequence of action chunks (each containing 8 actions), with the DRIS updated between chunks via SAM2 perception. 

In the \textbf{cluttered push-picking} task shown in Fig.~\ref{fig:realrobot} (left), the robot first pushed the target out of obstructing objects to create clearance, then grasped the target; the caging constraint kept the target within the DRIS distribution throughout the maneuver. In the \textbf{scooping} task in Fig.~\ref{fig:realrobot} (right), a flat object lying flush on the table could not be grasped directly from above, so the robot pushed it toward a wall corner and exploited the environmental geometry to scoop it up. This further confirms that the executor independence described in Sec.~\ref{sec:loop} enables sim-to-real transfer without fine-tuning.

\begin{figure*}[t]
  \centering
  \includegraphics[width=1\columnwidth]{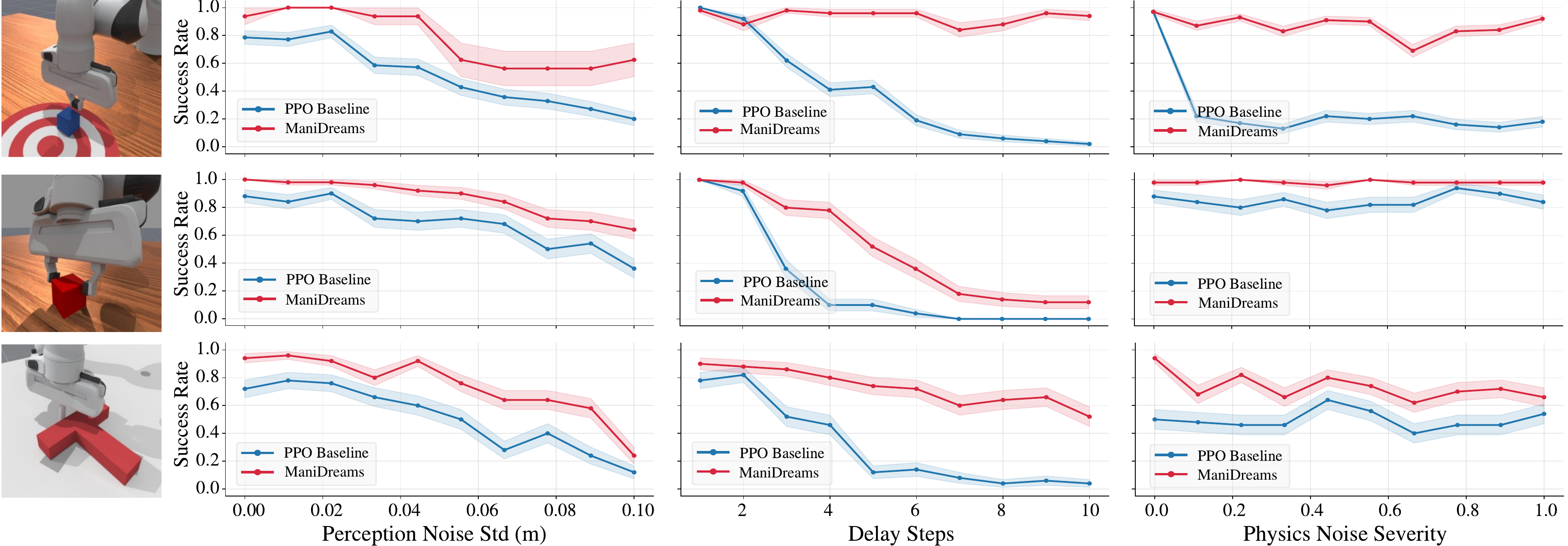}
  \caption{Success rate under increasing perturbations across three ManiSkill tasks (top to bottom: \textit{PushCube}, \textit{PickCube}, \textit{PushT}). \textit{Left}: Observation noise ($\sigma$ in meters). \textit{Center}: Observation delay (in simulation steps). \textit{Right}: Physics perturbation severity. Shaded regions indicate one standard deviation over 10 runs.}
  \label{fig:robustness}
  \vspace{-10pt}
\end{figure*}

\subsection{Robustness under Uncertainty}
\label{sec:exp_robust}

We compared the PPO baseline against ManiDreams framework under three perturbation types to evaluate robustness under perceptual noise, observation delay, and parametric uncertainty. The baseline executed the PPO policy directly through ManiDreams with \api{num\_samples=1} and no cage. ManiDreams used the same PPO policy as the sampler within the full sample-predict-constrain loop, with an $N$-best selector as the optimizer (Sec.~\ref{sec:loop}).

We evaluated on three ManiSkill3 default tasks with different characteristics: \textit{PushCube} (short-horizon, contact-sensitive), \textit{PickCube} (vertical grasping), and \textit{PushT} (longer-horizon, continuous pushing). For each perturbation type (observation noise, observation delay, and physics parameter randomization scale), we swept its magnitude as shown in Fig.~\ref{fig:robustness} while holding the other two at zero. Each configuration was repeated 10 times with 100 episodes per run; we report mean success rate and one standard deviation. ManiDreams used $N{=}8$ solver samples, $m{=}8$ DRIS instances, and caging weight $\lambda{=}0.1$ throughout.

As shown in Fig.~\ref{fig:robustness}, ManiDreams generally outperformed the PPO baseline across all nine task-perturbation combinations, with the advantage most pronounced at moderate perturbation levels. At extreme magnitudes, the fixed DRIS configuration can no longer fully cover the uncertainty, and both methods degrade together, indicating that adaptive DRIS configuration would be needed beyond this range. The degradation pattern is task-dependent: \textit{PickCube} involves direct grasping with minimal contact, making it resilient to physics variations but sensitive to observation delay; \textit{PushCube} is a short-horizon contact-rich task susceptible to friction and mass changes; \textit{PushT} requires sustained contact over a longer horizon, tolerating physics perturbations but vulnerable to stale observations that accumulate error over time.

Across all cases, the sample-predict-constrain loop provides robustness by evaluating candidates against distributional predictions: when uncertainty is mild, the filtering is largely redundant; when uncertainty is high, it becomes the critical difference.


\subsection{Ablation Study}
\label{sec:ablation}

We further ablated three key parameters on \textit{PushCube}: the number of DRIS instances $m$, the number of solver samples $N$, and the DRIS physics distribution width. To isolate the effect of each parameter, we fixed observation noise at $\sigma{=}0.04$\,m, action chunk length at 3, and physics perturbation severity at 0.4, then swept one parameter while holding the others at their defaults ($m{=}8$, $N{=}8$, Medium). Each configuration was evaluated over 100 episodes.

Table~\ref{tab:ablation} summarizes the results. Setting $m{=}1$ removes distributional information and degrades to point estimation; performance improves with more instances but saturates beyond $m{=}16$. Setting $N{=}1$ eliminates candidate comparison and reduces to direct policy execution; gains are consistent up to $N{=}16$ with diminishing returns thereafter. Distribution width, which scales default physics parameters by $[0.9, 1.2]\times$ (Narrow), $[0.5, 2]\times$ (Medium), and $[0.2, 3]\times$ (Wide), exhibits an inverted-U pattern: Narrow underestimates real variability (74\%), Wide introduces excessive conservatism (76\%), and Medium achieves the best balance (82\%).

\begin{table}[t]
\centering
\caption{Ablation study on \textit{PushCube}. Success rate (\%)
averaged over 100 episodes under combined perceptual and parametric
perturbations. Default configuration: $m\!=\!8$, $N\!=\!8$, Medium.}
\label{tab:ablation}
\vspace{2pt}
\renewcommand{\arraystretch}{1}

\begin{tabular*}{\columnwidth}{@{\extracolsep{\fill}} l | c c c c c}
    \hline
    DRIS instances $m$ & 1 & 4 & \textbf{8} & 16 & 32 \\
    Success rate(\%)    & 58 & 72 & 82 & 85 & 86 \\
    \hline
    Solver samples $N$ & 1 & 4 & \textbf{8} & 16 & 32 \\
    Success rate(\%)       & 52 & 71 & 82 & 87 & 88 \\
    \hline
    Distribution width & \multicolumn{2}{c}{Narrow} & \textbf{Medium} & \multicolumn{2}{c}{Wide} \\
    Success rate(\%)       & \multicolumn{2}{c}{74} & 82 & \multicolumn{2}{c}{76} \\
    \hline
\end{tabular*}
\end{table}


\subsection{Runtime Overhead}
\label{sec:exp_overhead}

\begin{figure}[t]
  \centering
  \includegraphics[width=0.6\columnwidth]{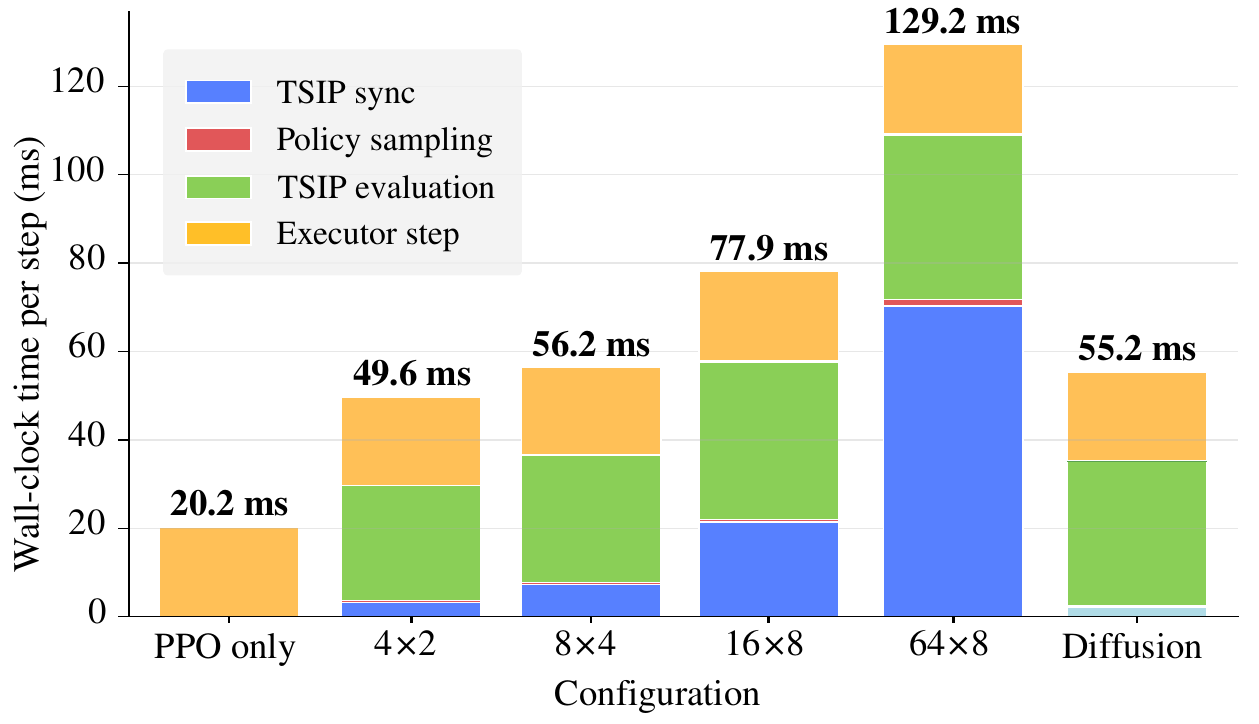}
  \caption{Wall-clock time per control step on \textit{PushCube} for increasing $N{\times}m$ configurations and a diffusion-based TSIP. }
  \label{fig:overhead}
  \vspace{-15pt}
\end{figure}

Finally, we evaluated the computational cost of the sample-predict-constrain loop on \textit{PushCube} while varying $N$ and $m$, as shown in Fig.~\ref{fig:overhead}. Baseline PPO ran at 20.2\,ms per step. Adding the sample-predict-constrain loop raised this to 49.6\,ms ($4{\times}2$), 56.2\,ms ($8{\times}4$), and 77.9\,ms ($16{\times}8$); the dominant costs were TSIP state synchronization and forward evaluation, while policy sampling and cage evaluation were negligible. The diffusion-based TSIP ran at 55.2\,ms, comparable to the simulation-based $8{\times}4$ configuration. Although all configurations exceed the 50\,Hz cycle, limiting applications for highly dynamic tasks, they remain practical for typical manipulation tasks at 10--20\,Hz, and in plan-then-execute mode per-step latency does not constrain the physical control rate.

\section{Conclusion}
\label{sec:conclusion}

We have presented ManiDreams, an open-source framework that unifies uncertainty representation, propagation, and constraint into a single composable pipeline for robotic manipulation. The core contribution of this work is not a state-of-the-art algorithm for any single task, but a modular, accessible platform that makes uncertainty-aware manipulation practical with reusable building blocks.

Experiments across three ManiSkill benchmark tasks, qualitative examples spanning different TSIP backends, solvers, and caging constraints, and real-robot deployment confirm that the framework adds robustness under perceptual, parametric, and structural perturbations, with negligible overhead over baseline policy execution and ease of integration. 

Current limitations include manual specifications of DRIS parameters and cages that require task-specific tuning. The framework also depends on the availability of a suitable TSIP backend; tasks with highly discontinuous dynamics (e.g., high-speed impacts) may require more expressive propagation models. Future work will explore automatic DRIS configuration from visual observations, integration with vision-language models for task-level cage specification, and extension to latent world models for wider applications.

\section*{Acknowledgment}
This work is supported by NSF-2240040.

\bibliographystyle{plainnat}
\bibliography{reference}

@string{ICRA = "IEEE International Conference on Robotics and Automation (ICRA)"}

@string{IROS = "IEEE International Conference on Intelligent Robots and Systems (IROS)"}

@string{IJRR = "The International Journal of Robotics Research"}

@string{RSS = "Robotics: Science and Systems"}

@string{RAL = "IEEE Robotics and Automation Letters"}

@string{RAM = "IEEE Robotics and Automation Magazine"}

@string{CoRL = "Conference on Robot Learning (CoRL)"}

@string{NeurIPS = "Conference on Neural Information Processing Systems (NeurIPS)"}

@string{ICLR = "International Conference on Learning Representations (ICLR)"}

@conference{tao2024maniskill3, 
    AUTHOR    = {Stone Tao AND Fanbo Xiang AND Arth Shukla AND Yuzhe Qin AND Xander Hinrichsen AND Xiaodi Yuan AND Chen Bao AND Xinsong Lin AND Yulin Liu AND Tse-Kai Chan AND Yuan Gao AND Xuanlin Li AND Tongzhou Mu AND Nan Xiao AND Arnav Gurha AND Viswesh Nagaswamy Rajesh AND Yong Woo Choi AND Yen-Ru Chen AND Zhiao Huang AND Roberto Calandra AND Rui Chen AND Shan Luo AND Hao Su}, 
    TITLE     = {{Demonstrating GPU Parallelized Robot Simulation and Rendering for Generalizable Embodied AI with ManiSkill3}}, 
    BOOKTITLE = RSS, 
    YEAR      = {2025},  
}

@article{cosmospolicy,
      title={Cosmos Policy: Fine-Tuning Video Models for Visuomotor Control and Planning}, 
      author={Moo Jin Kim and Yihuai Gao and Tsung-Yi Lin and Yen-Chen Lin and Yunhao Ge and Grace Lam and Percy Liang and Shuran Song and Ming-Yu Liu and Chelsea Finn and Jinwei Gu},
      year={2026},
      journal = {arXiv preprint arXiv:2601.16163},
}

@InProceedings{pmlr-v155-pinneri21a,
  title = 	 {Sample-efficient Cross-Entropy Method for Real-time Planning},
  author =       {Pinneri, Cristina and Sawant, Shambhuraj and Blaes, Sebastian and Achterhold, Jan and Stueckler, Joerg and Rolinek, Michal and Martius, Georg},
  booktitle = CoRL,
  year = 	 {2020}
}

@ARTICLE{cageinmotion,
  author  = {{Dong}, Yifei and {Cheng}, Xianyi and {Pokorny}, Florian T.},
  journal = RAL,
  title   = {Characterizing manipulation robustness through energy margin and caging analysis},
  year    = {2024},
  volume  = {9},
  number  = {9},
}

@article{MPPI,
    author = {Williams, Grady and Aldrich, Andrew and Theodorou, Evangelos A.},
    title = {Model Predictive Path Integral Control: From Theory to Parallel Computation},
    journal = {Journal of Guidance, Control, and Dynamics},
    volume = {40},
    number = {2},
    year = {2017},
}

@article{hafner2023dreamerv3,
  author  = {{Hafner}, Danijar and {Pasukonis}, Jurgis and {Ba}, Jimmy and {Lillicrap}, Timothy},
  title   = {Mastering diverse control tasks through world models},
  journal = {Nature},
  year    = {2025},
  volume  = {640}
}

@conference{hansen2024tdmpc2,
  author  = {{Hansen}, Nicklas and {Su}, Hao and {Wang}, Xiaolong},
  title   = {{TD}-{MPC}2: Scalable, robust world models for continuous control},
  year    = {2024},
  booktitle = ICLR,
}

@conference{tobin2017domain,
  author       = {{Tobin}, Josh and {Fong}, Rachel and {Ray}, Alex and {Schneider}, Jonas and {Zaremba}, Wojciech and {Abbeel}, Pieter},
  title        = {Domain randomization for transferring deep neural networks from simulation to the real world},
  year         = {2017},
  booktitle    = IROS,
}

@conference{peng2018sim,
  author       = {{Peng}, Xue Bin and {Andrychowicz}, Marcin and {Zaremba}, Wojciech and {Abbeel}, Pieter},
  title        = {Sim-to-real transfer of robotic control with dynamics randomization},
  year         = {2018},
  booktitle    = ICRA,
}

@conference{dong2024codesigningtoolscontrolpolicies,
  author       = {{Dong}, Yifei and {Han}, Shaohang and {Cheng}, Xianyi and {Friedl}, Werner and {Cabral Muchacho}, Rafael I. and {Roa}, M{\'a}ximo A. and {Tumova}, Jana and {Pokorny}, Florian T.},
  title        = {CageCoOpt: Enhancing Manipulation Robustness through Caging-Guided Morphology and Policy Co-Optimization},
  year         = {2025},
  booktitle = IROS
}

@article{muratore2022robot,
  author  = {{Muratore}, Fabio and {Ramos}, Fabio and {Turk}, Greg and {Yu}, Wenhao and {Gienger}, Michael and {Peters}, Jan},
  title   = {Robot learning from randomized simulations: A review},
  year    = {2022},
  journal = {Frontiers in Robotics and AI},
}

@article{yang2024diffusion,
  author  = {{Yang}, Ling and {Zhang}, Zhilong and {Song}, Yang and {Hong}, Shenda and {Xu}, Runsheng and {Zhao}, Yue and {Zhang}, Wentao and {Cui}, Bin and {Yang}, Ming-Hsuan},
  title   = {Diffusion models: A comprehensive survey of methods and applications},
  journal = {{ACM} Computing Surveys},
  year    = {2023},
  volume  = {56},
}

@article{rimon1999caging,
  author  = {{Rimon}, Elon and {Blake}, Andrew},
  title   = {Caging planar bodies by one-parameter two-fingered gripping systems},
  journal = IJRR,
  year    = {1999},
  volume  = {18},
  number  = {3},
  pages   = {299--318},
}

@article{rodriguez2012caging,
  author  = {{Rodriguez}, Alberto and {Mason}, Matthew T. and {Ferry}, Steve},
  title   = {From caging to grasping},
  journal = IJRR,
  year    = {2012},
  volume  = {31},
  number  = {7},
}

@article{makita2017survey,
  author  = {{Makita}, Satoshi and {Wan}, Weiwei},
  title   = {A survey of robotic caging and its applications},
  journal = {Advanced Robotics},
  year    = {2017},
  volume  = {31},
  number  = {19},
  pages   = {1071--1085},
}

@article{wang2025cagingintime,
  author  = {{Wang}, Gaotian and {Ren}, Kejia and {Morgan}, Andrew S. and {Hang}, Kaiyu},
  title   = {Caging in time: A framework for robust object manipulation under uncertainties and limited robot perception},
  year    = {2025},
  journal = IJRR,
}

@article{sucan2012ompl,
  author  = {Sucan, Ioan A. and Moll, Mark and Kavraki, Lydia E.},
  title   = {The {O}pen {M}otion {P}lanning {L}ibrary},
  journal = RAM,
  year    = {2012},
}

@inproceedings{makoviychuk2021isaacgym,
  author    = {{Makoviychuk}, Viktor and {Wawrzyniak}, Lukasz and {Guo}, Yunrong and {Lu}, Michelle and {Storey}, Kier and {Macklin}, Miles and {Hoeller}, David and {Ruber}, Norman and {Allshire}, Arthur and {Handa}, Ankur and {State}, Gavriel},
  title     = {{Isaac Gym}: High performance {GPU}-based physics simulation for robot learning},
  booktitle = NeurIPS,
  year      = {2021},
}

@inproceedings{wu2022daydreamer,
  author    = {{Wu}, Philipp and {Escontrela}, Alejandro and {Hafner}, Danijar and {Abbeel}, Pieter and {Goldberg}, Ken},
  title     = {{DayDreamer}: World models for physical robot learning},
  booktitle = CoRL,
  year      = {2022},
}

@article{ding2024diffusionwm,
  author  = {Zihan Ding and Amy Zhang and Yuandong Tian and Qinqing Zheng},
  title   = {Diffusion World Model: Future Modeling Beyond Step-by-Step Rollout for Offline Reinforcement Learning},
  journal = {arXiv preprint arXiv:2402.03570},
  year    = {2024},
}

@article{calli2017ycb,
  title={Yale-CMU-Berkeley dataset for robotic manipulation research},
  author={Calli, Berk and Singh, Arjun and Bruce, James and Walsman, Aaron and Konolige, Kurt and Srinivasa, Siddhartha and Abbeel, Pieter and Dollar, Aaron M},
  journal=IJRR,
  year={2017},

}

@conference{alonso2024diffusion,
  author       = {{Alonso}, Eloi and {Jelley}, Adam and {Micheli}, Vincent and {Kanervisto}, Anssi and {Storkey}, Amos and {Pearce}, Tim and {Fleuret}, Fran{\c{c}}ois},
  title        = {Diffusion for world modeling: Visual details matter in Atari},
  year         = {2024},
  booktitle=NeurIPS
}

@conference{ravi2024sam2,
  title={SAM 2: Segment Anything in Images and Videos},
  author={Ravi, Nikhila and Gabeur, Valentin and Hu, Yuan-Ting and Hu, Ronghang and Ryali, Chaitanya and Ma, Tengyu and Khedr, Haitham and R{\"a}dle, Roman and Rolland, Chloe and Gustafson, Laura and Mintun, Eric and Pan, Junting and Alwala, Kalyan Vasudev and Carion, Nicolas and Wu, Chao-Yuan and Girshick, Ross and Doll{\'a}r, Piotr and Feichtenhofer, Christoph},
  booktitle=ICLR,
  year={2025}
}

@inproceedings{chebotar2018closing,
  author    = {{Chebotar}, Yevgen and {Handa}, Ankur and {Makoviychuk}, Viktor and {Macklin}, Miles and {Issac}, Jan and {Ratliff}, Nathan and {Fox}, Dieter},
  title     = {Closing the sim-to-real loop: Adapting simulation randomization with real world experience},
  booktitle = ICRA,
  year      = {2019},
}

@inproceedings{mehta2019active,
  author    = {{Mehta}, Bhairav and {Diaz}, Manfred and {Golemo}, Florian and {Pal}, Christopher J. and {Paull}, Liam},
  title     = {Active domain randomization},
  booktitle = CoRL,
  year      = {2019},
}

@inproceedings{oscbf2025,
  author    = {Daniel Morton and Marco Pavone},
  title     = {Safe, Task-Consistent Manipulation with Operational Space Control Barrier Functions},
  booktitle = IROS,
  year      = {2025},
}

@article{reviewdynamics,
author = {Bo Ai  and Stephen Tian  and Haochen Shi  and Yixuan Wang  and Tobias Pfaff  and Cheston Tan  and Henrik I. Christensen  and Hao Su  and Jiajun Wu  and Yunzhu Li },
title = {A review of learning-based dynamics models for robotic manipulation},
journal = {Science Robotics},
volume = {10},
number = {106},
year = {2025},
}

@article{nudge2026,
  author  = {Haixin Jin and Nikhil Uday Shinde and Soofiyan Atar and Hongzhan Yu and Dylan Hirsch and Sicun Gao and Michael C. Yip and Sylvia Herbert},
  title   = {Learning to nudge: A scalable barrier function framework for safe robot interaction in dense clutter},
  journal = {arXiv preprint arXiv:2601.02686},
  year    = {2026},
}

@article{mayne2005robustmpc,
  author  = {{Mayne}, David Q. and {Seron}, Mar{\'\i}a M. and {Rakovi{\'c}}, Sa{\v{s}}a V.},
  title   = {Robust model predictive control of constrained linear systems with bounded disturbances},
  journal = {Automatica},
  volume  = {41},
  number  = {2},
  pages   = {219--224},
  year    = {2005},
}

\end{document}